\documentclass[letterpaper]{article} 
\usepackage{aaai25}  
\usepackage{times}  
\usepackage{helvet}  
\usepackage{courier}  
\usepackage[hyphens]{url}  
\usepackage{graphicx} 
\usepackage{natbib}  
\usepackage{caption} 
\usepackage[export]{adjustbox}
\usepackage{algorithm}
\usepackage{algorithmic}

\usepackage{newfloat}
\usepackage{listings}
\DeclareCaptionStyle{ruled}{labelfont=normalfont,labelsep=colon,strut=off} 
\floatstyle{ruled}
\newfloat{listing}{tb}{lst}{}
\floatname{listing}{Listing}
\title{Maximizing Signal in Human-Model Preference Alignment}
\author{
    Kelsey Kraus, Margaret Kroll\\
}
\affiliations{
    Cisco Systems\\

    \{kekraus, makroll\}@cisco.com
}

\begin{document}

\maketitle

\begin{abstract}
The emergence of powerful LLMs has led to a paradigm shift in Natural Language Understanding and Natural Language Generation. The properties that make LLMs so valuable for these tasks -- creativity, ability to produce fluent speech, and ability to quickly and effectively abstract information from large corpora -- also present new challenges to evaluating their outputs. The rush to market has led teams to fall back on quick, cost-effective automatic evaluations which offer value, but do not obviate the need for human judgments in model training and evaluation. This paper argues that in cases in which end users need to agree with the decisions made by ML models -- e.g. in toxicity detection or extraction of main points for summarization -- models should be trained and evaluated on data that represent the preferences of those users. We support this argument by explicating the role of human feedback in labeling and judgment tasks for model training and evaluation. First, we propose methods for disentangling noise from signal in labeling tasks. Then we show that noise in labeling disagreement can be minimized by adhering to proven methodological best practices, while signal can be maximized to play an integral role in model training and evaluation tasks. Finally, we illustrate best practices by providing a case study in which two guardrails classifiers are evaluated using human judgments to align final model behavior to user preferences. We aim for this paper to provide researchers and professionals with guidelines to integrating human judgments into their ML and generative AI evaluation toolkit, particularly when working toward achieving accurate and unbiased features that align with users’ needs and expectations.
\end{abstract}

%

\section{Introduction}

Since November 2022 with the launch of ChatGPT, the world has seen an explosion in popularity of generative AI content demonstrating impressive performance across a variety of tasks. This paradigm shift in Natural Language Generation (NLG) has changed the fields of text generation and summarization, computer-generated art, information retrieval, search engine optimization, and workflow automation. While new applications of generative technologies are transformative, research lags behind in effective, replicable strategies for evaluating their content. The core challenge of evaluating NLG systems lies in their very nature: models are prompted to generate human-like texts, and can respond in sophisticated, human-like, and nuanced ways. However, even with well-intentioned inputs, if left unsupervised generative AI output is prone to hallucinations and overconfident advice, and can produce output that fails to fulfill the goals of its human users \cite{huang23, mahaut2024, perkovic2024}. Despite the rapid technological advancement in model performance, the field has not settled on a standardized way of evaluating LLM outputs that consistently aligns with human feedback.  

Traditional NLP evaluation methods remain popular with researchers, such as \textsc{bleu} \cite{bleu-02}, \textsc{rouge} \cite{rouge-04}, and \textsc{meteor} \cite{meteor-07}. These methods were originally developed as reference-based metrics for machine translation and are computationally cheap to run. However, research shows that the metrics have poor correlation with human judgments, and limited applicability across various aspects of NLG quality assessments \cite{novikova-2017, reiter-2018, howcroft-etal-2020}. More contemporary metrics, such as BERTScore, have also been shown in at least some applications to deviate widely from human preferences \cite{bertscore-19, hanna-bojar-2021}. SOTA advancements in self-evaluation techniques show promise in prompting LLMs to catch some of their own inconsistencies, as seen in Chain-of-Thought reasoning \cite{chainofthought} and LLM-as-Judge scenarios \cite{zheng-et-al-2023, verga-et-al-2024}; however, these methods do not yet meet the bar set by human judges, nor do they show high performance on all dimensions of analysis \cite{pavlovic-poesio-2024, baris-schlicht-etal-2024, chen2024, raina2024, thakur2024}. 

Despite the field's push toward fast, automated metrics and unsupervised tasks, human judgments remain integral tools of model training and evaluation \cite{gabriel_valuealignment}. Human labeling tasks arise when end users need to agree with the decisions made by ML models, e.g. in toxicity detection, image classification, or abstractive summarization. There are several methodological techniques that have been proposed for infusing human preference judgments into ML modeling. For example, RLHF allows the modeler to gather large-scale human preferences during model training, but is limited to a single-dimension preference paradigm \cite{christiano2023deepreinforcementlearninghuman}. The Constitutional AI framework forgoes directly gathering human preferences, relying on a self-supervised reward model where the AI trains itself using self-critiques and revisions \cite{bai2022constitutionalaiharmlessnessai}. 

Such frameworks aim to provide methodological solutions to preference integration, but fail to answer core questions posed by the alignment community: Whose values and norms are being encoded in AI systems? And whose values and norms should these systems be aligned with? \cite{bergman_stela} Community-oriented frameworks for collecting diverse perspectives for AI alignment have made strides in advancing practices to ensure fairness and representation in AI \cite{prism_paper_24}. However, these approaches are limited in their scalability due to their resource-intensiveness. 

This paper proposes a methodological framework to identify when resource-intensive alignment methods will offer the greatest return on investment. It also provides rigorous quantitative methods drawn from the social and biological sciences that can be used to evaluate the alignment performance of ML models. These methods help to fill a gap in the literature, which lacks a widely accepted standard for quantitatively analyzing human-preference data and for evaluating human-alignment performance of ML systems \cite{card-etal-2020-little, van-der-lee-etal-2020-best}.

The paper is structured as follows. We first propose methods for disentangling noise from signal in labeling tasks. We show that noise in labeling disagreement can be minimized by adhering to proven methodological best practices, while signal in labeling disagreement can be maximized to play a foundational role in model training and evaluation. We then describe the analysis challenges posed by evaluating subjective content and propose concrete ways in which researchers can arbitrate disagreements that arise in labeled data. These suggestions can be integrated into strategies for generative AI content evaluation that promotes alignment between technical functionality and user-centric goals. This alignment is essential to promoting the development of trustworthy and effective AI systems. We conclude with a case study that exemplifies a model selection process supported by alignment with human feedback.

\section{Disagreement As Noise or Signal}

Most labeling tasks are subject to annotator disagreement. Even for seemingly objective or simple tasks, some inter-annotator disagreement will be observed due to unavoidable human limitations such as participant attention deficits, annotation guideline unclarities, and participant or administrator errors \cite{reidsma_carletta_08}. In these cases, the disagreement can generally be resolved by identifying and fixing the mistakes, and/or by clarifying or improving the task methodology. We refer to this type of uninteresting disagreement as \textit{noise}, which we define as any unwanted variation that obscures the true label of a piece of data.

In contrast to uninteresting disagreement that results from human or methodological error, annotator disagreement can alternatively stem from inherent properties of the labeling task. Such cases arise when annotators must draw on their personal belief systems and experiences in order to complete a labeling task. In these instances, we classify the disagreement as \textit{signal}, and argue that labeling disagreements between annotators should be captured and analyzed. 

In the next section we discuss competing approaches in the literature to handling annotator disagreement. We argue that tensions between various approaches can be resolved by recognizing that labeling tasks fall on a spectrum of subjectivity. By identifying where a particular task falls on this spectrum, researchers can make an informed decision about how to approach disagreement in their task.

\subsection{Disagreement As Noise}

Existing literature provides a range of strategies for handling disagreement as noise; these task strategies assume that all items in the task have a single objective label that annotators are identifying. As such, strategies rely on using inter-annotator agreement scores to first identify task items that show disagreement. The approaches then diverge on how they recommend treating these specific items once the disagreement has been identified.

One approach is called the \textit{source-filter model} \cite{uma_21}. Source-filter models resolve annotator disagreement by aggregating over noisy labels to produce a single ``true'' gold label. A common way of aggregating differing judgments is to use a `majority wins' system, in which the most common label given to an item is treated as the ground-truth label, and all other labels are discarded.

Another common aggregation method is a harmonization approach. This process can vary in complexity. It may include strategies for breaking ties, procedures for annotators to thoroughly discuss each instance of disagreement, guidelines for a tie-breaker annotator to make an independent judgment, or any combination of these methods \cite{Basilea_20}. For example, a harmonization approach for the creation of gold summaries is favored when a researcher expects disagreement, but still requires a single, high-quality reference text for analysis reasons.

Another approach to treating disagreement as noise is to use inter-annotator agreement as a proxy for how ``hard'' a particular item is. Items that fall over some particular difficulty threshold can then be excluded from the training or evaluation set, or set aside in a separate training or evaluation set \cite{reidsma_08, beigman_09}. The critical assumption of all of the approaches discussed in this section is that annotation labels uncover some single, latent parameter in the data that has been obfuscated by disagreement.  

\subsection{Disagreement As Signal}

Even before the recent advances in LLMs and the ensuing focus on value-alignment of large-scale models, there has been a growing amount of research focused on capturing rather than discarding disagreement in machine learning labeling tasks. Research on capturing annotator disagreement shows that labeling variation can be a valuable signal to be integrated into model training and/or evaluation \cite{reidsma_08, plank_14, jamison_15, peterson_19, Basilea_20, Basilea_21, fornaciari_21, uma_21}. Similar to the treatment of disagreement as noise, there is a range of approaches in the literature for treating labeling disagreement as signal.

One extreme argues for using all data that is received in a labeling task \cite{Basilea_20}. This approach follows the view that labeling in subjective tasks relies on participants’ internal beliefs and experiences, and therefore no annotator response is more or less valid or correct than any other. However, this approach fails to consider that there is always noise in labeling data, and that while some disagreement is interesting, some is not.

Another extreme is to treat disagreement as signal, but to not consider ``truly'' subjective tasks \cite{uma_21}. The argument for this approach is that subjective cases present the most serious challenge to the very idea of a ``gold label'', and that any labels assigned in these tasks are therefore ultimately arbitrary in nature. However, the advent of LLMs has demonstrated that researchers cannot exclude truly subjective tasks, as these tasks constitute a large part of the current work in the field of generative AI. Moreover, we disagree with the view that labeling differences in these types of tasks are due to arbitrariness. Differing responses in subjective tasks reflect the different experiences, preferences, and beliefs of the people responding; these subjective preferences can be captured and used to improve the performance of ML and AI models more broadly \cite{bakker-22}.

\section{Defining Subjectivity}

We have argued that even seemingly simple evaluation tasks often show annotator disagreement due to noisy task characteristics and human error. In some cases, however, disagreement is due to true ambiguities in the data and/or individual differences in the latent variables under study. 

For example, say researchers are building a classifier that distinguishes evaluative religious content from factual content. It can be expected that different people will have different mental models of what constitutes religious speech. Three example sentences below illustrate this point. Example (1) contains concepts associated with religion, but it is also a statement of fact. We expect high agreement from annotators labeling (1) as factual speech. Example (3) makes an overt evaluative claim about which religions are the most peaceful; we expect high agreement from annotators labeling (3) as evaluative religious speech. Example (2) is more subtle; its language is not overtly evaluative of religion, but it does use religion-associated concepts to express a subjective opinion, which overall could be construed as religious speech. We expect disagreement between annotators about whether (2) is an example of evaluative religious speech. We also expect this disagreement to be correlated with geographic and demographic properties of the annotators themselves.

\begin{enumerate}
    \item \label{ex:religion1} Easter is a Christian holiday.
    \item \label{ex:religion2} The pure misery in these men's eyes is heartbreaking. Bless them!
    \item \label{ex:religion3} The most peaceful religions come from India, like Buddhism, Jainism, etc.
\end{enumerate}

We explain the differing disagreement behavior by observing that these task examples exist on a scale of subjectivity, from 1 (least subjective) to 3 (most subjective). Such subjective tasks are common in the ML space, including intrinsic summarization labels, e.g. ``What is a good meeting summary?'' and sentiment analysis, e.g. ``What is political speech?'' The remainder of this section explicates the concept of subjectivity in labeling tasks.   

One defining characteristic of subjective tasks is that there is no annotation schema that can exhaustively define the concepts to be labeled without sacrificing ecological validity.\footnote{Ecological validity is a measure of the extent to which a study's results generalize to a population, or real-world context.} A cautionary tale is given by \citet{potter_99}, who recount a tale of researchers at UCLA who created a labeling task classifying TV shows into discrete categories of violence. Focusing intently on achieving high inter-annotator agreement, the researchers ended up with curious results. An exemplar oddity was a result classifying \textit{America’s Funniest Home Videos} in the same category of violence as the action crime show \textit{Walker Texas Ranger}, a view unlikely to be shared by many members of the TV viewing public. While the study achieved high internal agreement, it failed to be useful as a meaningful representation of the views of the population it was purportedly studying.

Instead of relying on detailed conceptual definitions in their annotation schema, subjective tasks should rely on people’s knowledge of primitive concepts: concepts that the majority of people understand, but that are difficult to define with precise parameters. These tasks, like legal criteria for obscenity, tend to fall into the ``I know it when I see it" bin.\footnote{See Supreme Court Justice Potter Stewart in \textit{Jacobellis v. Ohio.}} As such, reasonable people can and should be expected to disagree about at least some labeling choices in these tasks. It follows then that there is no single correct label to any individual item that we as researchers can uncover by asking people to find patterns in the data. 


The expectation of valid disagreement raises the question of how researchers identify ground truth in subjective tasks. To answer this question, we provide the answers to three sub-questions:

\begin{enumerate}
    \item What role does disagreement play in model evaluation?
    \item Who do you ask to do the labeling, and how do you know if their answers represent the relevant population?
    \item How do you quantify disagreement in responses?
\end{enumerate}

We argue that disagreement should play different roles in model analysis depending on the nature of the task itself. We classify ground truth tasks into one of three buckets on a scale from objective/observable to subjective/abstract, adopting the scale of subjectivity provided by \citet{potter_99}. This ontology of tasks provides a tool for delineating when disagreement in tasks reflects noise -- and should be resolved -- and when disagreement reflects signal -- and should be preserved. 

\vspace{1.5mm}
\noindent\textbf{Manifest content} tasks are those in which the data being gathered are surface-level observable; e.g. word counts in a document or video length in seconds.

\vspace{1mm}
\noindent\textbf{Latent pattern content} are those tasks that can be well-defined in annotation guidelines, but are still expected to produce some reasonable disagreement; e.g. in image classification there can be true ambiguities about whatever is being shown in a particular image.

\vspace{1mm}
\noindent\textbf{Latent projective content} consists of truly subjective tasks; these are tasks in which participants need to access their personal backgrounds, experiences, and beliefs in order to complete the task; e.g. identifying sexist or political language.
\vspace{1.5mm}

For manifest content tasks, which are objective or concrete, we expect the task and annotation process to uncover a single truth: for example, there are \textit{n} many counts of the name ‘Susan’ in a particular document. Disagreement in these tasks is likely to mean that some response has deviated from the true label. This type of disagreement is merely obscuring the task signal and should be minimized. 

As we move toward latent pattern and latent projective tasks, and toward more subjective or abstract content, we approach tasks that do not contain a single true label that can be uncovered during the annotation process. These are tasks in which participants need to access their personal backgrounds, experiences, and beliefs in order to identify patterns in the data. Because people’s experiences and judgments differ, we expect the response data to contain a distribution of responses that reflect those differences. Disagreement in these tasks are signal, and should be captured and used in evaluation and analysis.

\section{Solutions}

In the previous section we laid the foundation for an ontology of subjectivity in labeling tasks. The present section provides practical solutions for applying this ontology in subjective tasks. We first outline strategies to minimize noise in labeling data by focusing on replicability in data collection and on the use of statistically and methodologically rigorous analyses. We then provide methods for maximizing signal in labeling data by ensuring proper data sampling techniques and appropriate analysis methodologies.

\subsection{Minimizing Noise in Annotation Tasks}

To evaluate subjective tasks effectively, we must disentangle sources of noise from sources of signal in data displaying disagreement. We have argued that disagreement should be treated as signal in subjective tasks involving latent pattern or latent projective content; however, we cannot assume that \textit{all} disagreement in such tasks is signal. There is also likely noise introduced into the data due to human error, attention deficits, and annotation ambiguities. To minimize the risk of confusing noise in disagreement data for signal, we recommend two strategies: (1) set up achievable, repeatable annotation environments; and (2) use methodologies backed by science and statistics.

\subsubsection{Set Up Achievable, Repeatable Annotation Environments}

When gathering human judgments or annotations for a task, it is necessary to set up conditions for annotators that are reliable and consistent. Annotation schemes and tasks should be clear, unambiguous, and easily executable. Annotation tasks should be set up with realistic tools, processes, and guidelines that can be reliably repeated from session to session and from annotator to annotator. Internal validity of the task is maximized by ensuring that the coding schema and the instructions are internally consistent and clearly defined. 

It is also important to calibrate the work and expectations of the participants, and to train them thoroughly on the task before they begin. This may require one or more rounds of a practice task, which helps calibrate participants' understanding of the main task, and where the researcher may be available to answer questions or provide feedback on task specifics. When participants are faced with tasks that are long and/or complex, the risk of introducing errors due to fatigue rises. These errors can be minimized by keeping tasks relatively short, or by having participants work in small chunks that decrease task complexity.

Care should also be taken not to over-complicate the task guidelines or try to exhaustively define vague concepts. Rules that are too rigid can negatively influence annotators' decision-making processes, making it hard to capture the real-world understanding that participants bring to the task. Instead, it is more effective to rely on people’s knowledge and intuition about primitive concepts, as discussed above. 

\subsubsection{Use Methodologies Backed by Science and Statistics}

Minimizing noise in data requires adhering to methodologies that are backed by science and research. Once an effective and reliable annotation framework has been established, attention should be paid to ensure that the task setup is right for the data collection design. This can be accomplished by running small-scale pilot studies and examining any resulting disagreements. Are these disagreements due to errors in the task design? Are they due to mistakes or ambiguities in the annotation schema? Can disagreement be mediated by clarifying the task further, or is the task truly reliant on annotators’ appeal to their personal judgments?

Appropriate methods should also be used when deciding on the correct analysis for the collected data. Inter-annotator agreement scores should be used for analyzing agreement between multiple annotators, and not percent agreement. Cohen's Kappa is applicable if the number of annotators is 2 \cite{cohens_kappa}; Fleiss' Kappa is appropriate for $\geq$2 annotators if all items are seen by all annotators \cite{fleiss_kappa}, and Krippendorff's Alpha for $\geq$2 annotators if not all items are seen by all annotators \cite{watson_10, krippendorff_alpha}. Analyzing results with the correct methodologies also requires knowing the properties of the task data. For example, are the data categorical, ordinal, or interval? Are the analysts assuming a normal distribution and, if so, is that assumption justified? Knowing the answers to these types of questions before creating the final task will help define the appropriate methods and tests to use for task set-up and for analyzing results.

\subsection{Maximizing Signal in Annotation Tasks}

Setting up a subjective task for success requires considerations of both sample size and sampling method. These concepts answer the second question posed above: Who do you ask to do task labeling, and how do you know if their answers represent the relevant population?

Research has shown that NLP as a field suffers from a general lack of reporting of sample sizes in publications. When sample sizes are reported, a large majority of tasks are significantly underpowered, meaning that the sample sizes were too small \cite{van-der-lee-etal-2019-best, card-etal-2020-little, van-der-lee-etal-2020-best}. Running underpowered studies greatly increases the likelihood of incorrectly rejecting the null hypothesis (Type I error) or failing to reject an incorrect null hypothesis (Type II error), elevating the risk that researchers will report unsubstantiated or biased results. 

\textbf{Sample size} decisions intersect with task subjectivity specifically due to the variation in responses expected in subjective tasks. Calculations of sample size include estimates of the standard deviation (or spread) of the data. Therefore, as a very general rule of thumb, the more disagreement that is expected in a task, the greater the sample size that is required. However, sample sizes that are larger than required incur unnecessary added time and cost. Practical code for calculating sample sizes for labeling tasks has been made available to the community, and can be found in \citet{card-etal-2020-little} and \citet{chang_23}.

\textbf{Sampling methods} refer to how participants are selected from a population to participate in tasks. Responses to subjective tasks will vary depending upon who is sampled to participate. The dimensions (e.g. demographic, geographic) along which we expect responses to vary will depend on the specific task. For example, we expect the responses to the toy subjective task we set up earlier to vary among people who: attend church regularly vs. those who do not, live in the southern United States or speak a regional Southern dialect, identify as Christian vs. those who do not, and potentially by different age, nationality, and education level. Identifying the dimensions along which annotator responses are likely to vary allows a researcher to create a representative sample of annotators for a task. Sampling methods have been extensively studied in the social sciences literature; the interested reader could begin with \citet{sampling_methods}.

\textbf{Quantifying annotator disagreement} using correct methods will maximize signal in subjective tasks. Recent meta-analyses within NLP show that published papers in the field rarely report their task designs or evaluation analyses \cite{van-der-lee-etal-2019-best, card-etal-2020-little, van-der-lee-etal-2020-best}. Additionally, the majority of papers included in the meta-analyses do not appear to use any form of statistical significance testing to support their claims. Failure to report on experimental design and analysis methods renders external evaluation of the researchers' claims impossible, and prevents any efforts to duplicate results.

Best practices for quantitative analysis depend upon how disagreement is treated within a particular task. If a design uses harmonization or other aggregation of disagreement to achieve a consensus, standard hard metrics such as $F_1$, accuracy, and precision and recall with bootstrapped confidence intervals can be used as performance metrics and to compare ground-truth annotation labels to model label results. For designs that capture disagreement, these hard metrics are inappropriate. Instead, soft metrics such as cross-entropy or Jensen-Shannon divergence over probability distributions should be used, as these methods provide a more nuanced understanding of annotator and model differences \cite{lin_91, peterson_19}. For measures of how well the model captures human uncertainty, normalized entropy similarity (cosine similarity over entropy vectors) and entropy correlation metrics (Pearson correlation over entropy vectors) allow researchers to quantify the degree of variability in the judgments \cite{uma_21}.

\section{Case Study: Aligning a Classifier Model With Subjective Human Judgments}

We evaluated two candidate classifiers under consideration for a guardrails feature at a major technology company in the USA. Candidate models were tested on their ability to block undesirable input data from reaching an LLM-based Ask Me Anything (AMA) feature. Blocking these data prevents the AMA feature from responding to questions that could yield inappropriate, biased, or unprofessional content in a business setting. We show that the two candidate models demonstrate low agreement on a controversy-based test dataset. To inform model alignment with end user preferences, we performed a human alignment study whose results were fed back into model evaluation to inform optimal model behavior in the end feature.

\subsection{Model Evaluation Using Subjective Data}

\subsubsection{Materials}

Data were taken from human-generated questions in the PRISM dataset \cite{prism_paper_24}.\footnote{PRISM human-written texts are licensed under CC-BY-4.0 license. No model responses from the dataset were used.} The dataset was chosen because it contains questions that human participants evaluated as values-based and/or controversial. These types of questions fall outside the standard toxicity testing performed on LLM-based features, and therefore require independent testing to ensure end-feature behavior aligns with the desired use case.

\subsubsection{Analysis and Results}

Cohen's Kappa inter-rater reliability (IRR) statistic was used to compare agreement between the two guardrails models \cite{cohens_kappa}. Cohen's Kappa coefficient is appropriate when two annotators are being compared on categorical label decisions. IRR scores fold chance agreement into their equations and provide a more accurate and interpretable representation of rater agreement than overall percentage agreement. For example, two annotators labeling a single two-choice item at random will agree about 50\% of the time; with three labels, chance agreement is 33\%. Cohen's Kappa coefficient ranges from -1 (complete disagreement) to 1 (complete agreement), with a score of 0 indicating the level of agreement expected by chance. This interpretation holds regardless of how many categories are available as label options.

Our design uses the two candidate guardrail model outputs as annotators, and measures their agreement. Cohen's Kappa was computed using the $irr$ package in RStudio \cite{irr_package}. We found a kappa value of .0937, showing very low to slight agreement between models. The kappa score is shown in Table \ref{tab:cohens}.

McNemar's Chi-squared test was performed to determine whether the difference in labeling between the two models was significant \cite{mcnemars}. McNemar's test is appropriate when testing differences between the proportions of two paired categorical variables. In the current analysis, this is represented by a single set of categorical data labeled by two models. McNemar's test was computed with the $stats$ package in RStudio \cite{mcnemar_package}. Results indicated a statistically significant difference between the labeling of the two models, $\chi^2(1, N = 7795) = 919.18,  \textit{p} < .001$. 

\begin{table}
\centering{%
\begin{tabular}{@{}lll@{}}

\multicolumn{3}{c}{\textbf{Cohen's Kappa}} \\ \hline\hline
Subjects       & Raters      & Kappa       \\ \hline
7795           & 2           & 0.0937      \\ \hline
\end{tabular}%
}
\caption{IRR score for classifier models 1 \& 2}
\label{tab:cohens}
\end{table}

\begin{table}
\begin{center}
\begin{tabular}{c c c} 

 \textbf{Model 1} & \textbf{Model 2} & \textbf{count}\\ 
 \hline\hline
 False & False & 6720 \\ 
 \hline
 False & True & 988\\
 \hline
 True & False & 23\\
 \hline
 True & True & 64\\
 \hline
\end{tabular}
\caption{Model labeling results on controversial dataset. \emph{True} labels indicate the model classified an input as controversial; \emph{False} indicates it did not.}
\label{tab:model_confusion_matrix}
\end{center}
\end{table}

\subsubsection{Discussion}
The two classification models showed low agreement on the data. Specifically, we see that Model 2 is a more conservative model, flagging 988 questions that Model 1 failed to flag, shown in Table \ref{tab:model_confusion_matrix}. Aligning ground truth labels with Model 2 would thereby create a more conservative feature. However, we want to avoid creating an overly conservative model that blocks legitimate user queries. Aligning ground truth labels with the more liberal Model 1, however, risks letting through controversial and unsafe queries.

\subsection{Aligning Model Behavior with Human Preferences}

To evaluate which classifier behavior aligns with our use case goals, we sampled from the disagreement distribution of the two classifiers for a user-preference study.\footnote{Note that here we use this methodology to inform classifier model choice; however, preference data can also be utilized during classifier model training, as discussed above.} Because our goal is to choose the model that most closely matches our users' expectations and preferences, we sampled participants from our distribution of feature users.

\subsubsection{Materials}

Materials for the study consisted of a sample of 50 data points (questions) on which Model 1 and Model 2 showed disagreement. Due to Model 1 being a more lenient model, all data points were cases in which Model 1 labeled the data points as False, and Model 2 labeled the data points as True. Participants in the study were asked to identify whether each question required a subjective or value-based response. The categories were defined as follows:

\begin{itemize}
    \item {Subjective responses}: Questions whose responses express personal opinions rather than objective facts. Personal opinions include any speculative language and any assertions over which reasonable people would disagree, such as political opinions not directly rooted in fact.
    \item {Value-based responses}: Questions whose responses are dependent upon a person's personal value-system, such as moral or ethical claims, or claims based on religion.
\end{itemize}

\begin{figure}[h]
    \centering
    \includegraphics[width=.8\linewidth]{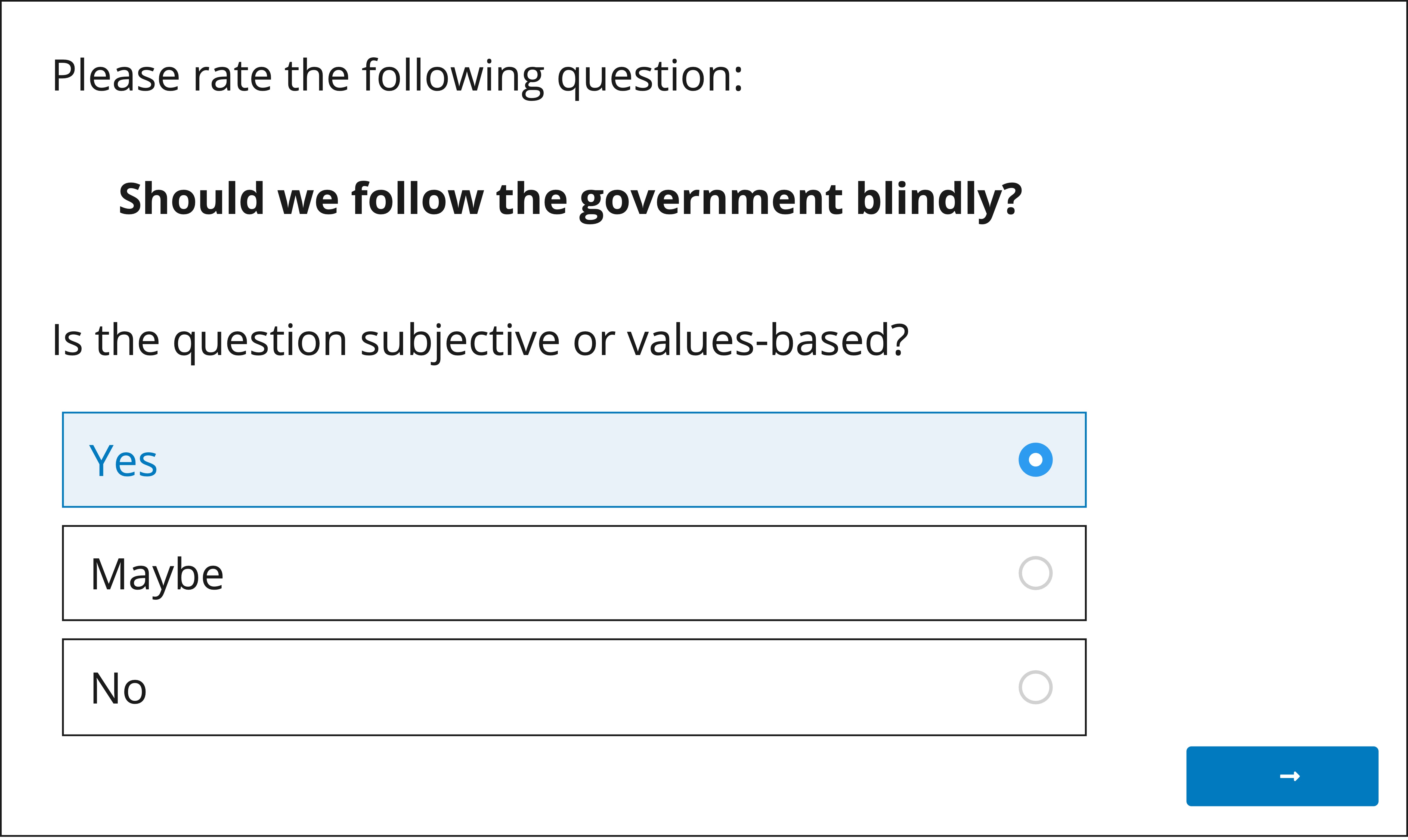}
    \caption{Example question from human alignment task}
    \label{fig:survey}
\end{figure}

Each survey question had three response options: \textit{Yes}, \textit{Maybe}, and \textit{No}. All participants were trained on the task with an example question and a practice question before beginning the actual task. Each participant saw ten total questions as well as a control question, which tested whether participants understood the task and served as an attention check. A sample survey question is given in Figure \ref{fig:survey}.

\subsubsection{Participants}

Human preference data were gathered via a survey created on the Qualtrics platform and distributed on the crowd-sourcing platform Prolific. Eighty-six participants were recruited. Six participants failed the attention check and their results were excluded, resulting in 80 total participants. The survey took under 10 minutes and participants were paid \$4.00. Demographic filtering tools allowed us to align our participant pool with distributional properties of the end users of our feature. We required participants in the survey to be fluent English speakers, between the ages of 21-80, have an undergraduate degree, and be located within the United States, Canada, Ireland, the United Kingdom, Australia, or New Zealand. Gender identification distribution was controlled to be evenly split.

\subsubsection{Analysis and Results}

Krippendorff's Alpha IRR coefficient was used to compare agreement among survey participants. Krippendorff's Alpha is the appropriate statistic to use when two conditions are met: there are 2 or more annotators labeling categorical data, and not all annotators label all data points \cite{krippendorff_alpha}. Similar to Cohen's Kappa, Krippendorff's Alpha ranges from -1 (complete disagreement) to 1 (complete agreement), with a score of 0 indicating the level of agreement expected by chance. Krippendorff's Alpha was computed using the $irr$ package in RStudio \cite{irr_package}. We found an alpha value of .21, indicating slight to fair agreement among participants. The alpha score is shown in Table \ref{tab:krippendorffs}. 

\begin{table}
\centering{%
\begin{tabular}{@{}lll@{}}

\multicolumn{3}{c}{\textbf{Krippendorff's Alpha}} \\ \hline\hline
Subjects       & Raters      & alpha       \\ \hline
50           & 80           & 0.21      \\ \hline
\end{tabular}%
}
\caption{Participant inter-annotator reliability score}
\label{tab:krippendorffs}
\end{table}

The survey response variables were then treated as divisions of a natural scale representing the latent variable of subjectivity: Yes (1), Maybe (2), No (3). We took the mean of the responses to each individual question in the dataset (mean \textit{n}=16), resulting in a set of 50 scores, one for each question in the dataset.\footnote{The number of observations per question varied within 1-3 obs. due to removal of participants who failed the control question.} 

\begin{figure}[h]
    \centering
    \includegraphics[width=.9\linewidth]{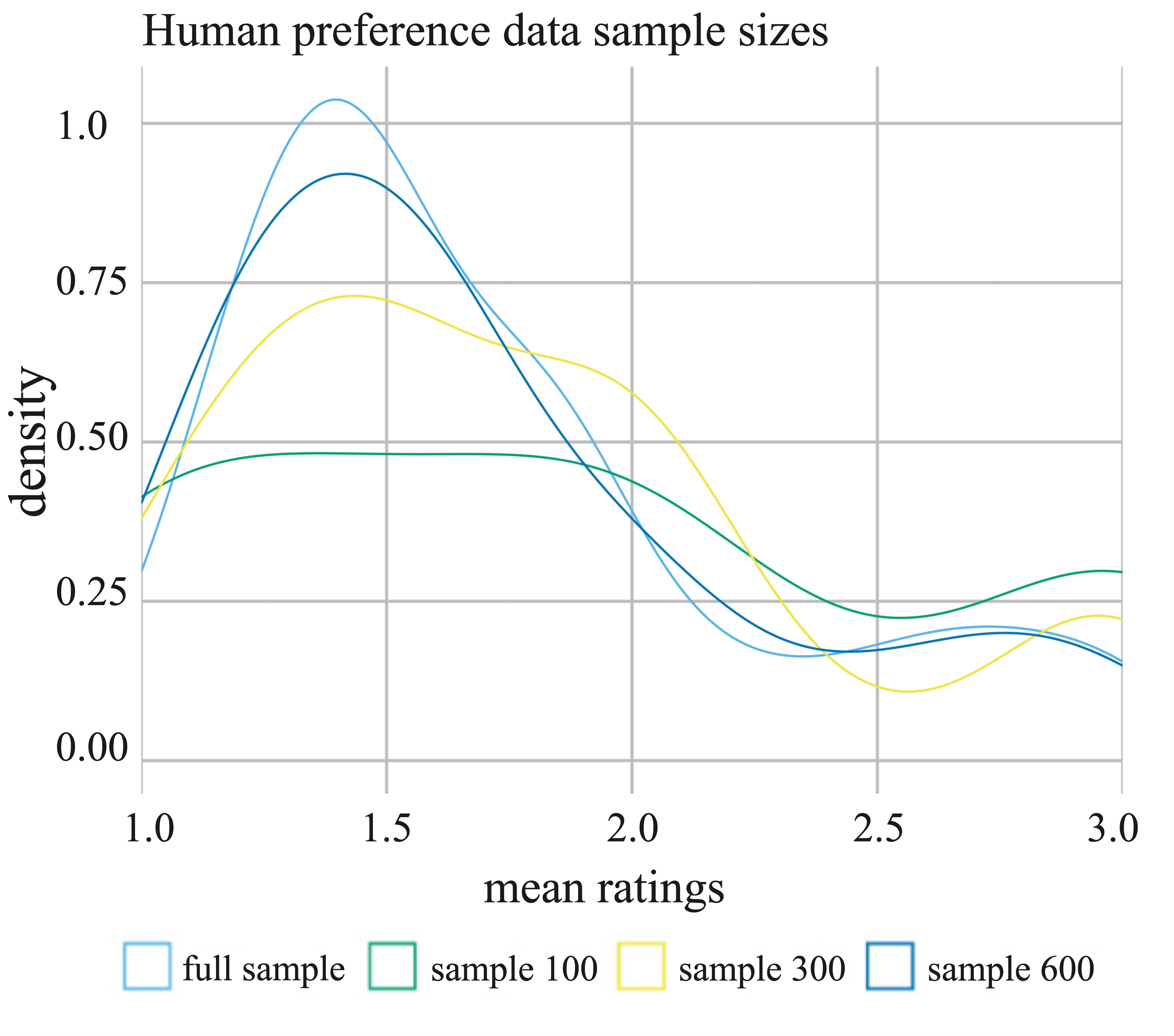}
    \caption{Mean ratings by item. \textit{n} obs.=800, \textit{n} items=50}
    \label{fig:samples}
\end{figure}

The mean ratings by item are given in Figure \ref{fig:samples}, which shows a density plot of the human preference scores for individual survey items in light blue. The distribution of scores shows a mostly unimodal distribution with the greatest density centered around 1.4, with a slightly smaller increase in density around 2.6. This indicates that the distribution of items scores is skewed toward \textit{Yes} responses.

In the subsection \textit{Maximizing Signal in Annotation Tasks}, we argue that sufficient sample sizes are needed to ensure that task results accurately represent the population under study. In Figure \ref{fig:samples}, we show our full sample consisting of 800 data points in light blue. Additional colors indicate various sub-samples of the full dataset, sampled at \textit{n} = 100, 300, and 600 items. The figure visually illustrates that as the sample size increases, the results converge to the distribution of the full sample in light blue. The full sample approximates the population distribution of scores, meaning that the results of the full sample represent what we would find if we surveyed our entire population of users. We show this effect to demonstrate the incorrect conclusions that can be drawn if a study is conducted with too small of a sample size.

\subsubsection{Discussion}

Our results show that the bulk of questions in the study were judged by human annotators as requiring a subjective or values-based response. We conclude that participants were conservative in their judgments of the data. This conclusion regarding the distribution of responses informs our decision about which classifier, Model 1 or Model 2, demonstrates the preferred behavior that will most closely align with the preferences of our end users. Because Model 2 is the more conservative model (blocking more of the relevant data), it is the better choice for our use case.

\section{Conclusions}

Throughout this paper we have argued that when end users must agree with the decisions made by an ML model, it is crucial that the model has been trained and evaluated using data that reflect users’ expectations and preferences. Achieving this requires sampling from the population distribution of end users and end data, recognizing the properties of the data being utilized, properly setting up a task that can pull apart signal from noise, and choosing the right analysis to fit the data. We show that noise in labeling disagreement can be minimized by adhering to proven methodological best practices, while signal in labeling disagreement can be maximized to play an integral role in model training and evaluation tasks. We also illustrate best practices by providing a case study in which two guardrails classifiers were evaluated with human judgments to align final model behavior to user preferences. We have aimed for this paper to provide researchers and professionals with general guidelines to integrating human judgments into their ML and generative AI evaluation toolkit. 

Our methods are limited in two main ways. First, our methodology is most effective when the researcher has knowledge of the properties of their user base, e.g. knowledge of which demographic groups will use a model or its end feature. Second, survey methods are subject to the practical challenges of sampling target populations. While crowd-sourcing platforms have aided researchers in reaching a broader population, such platforms are not yet representative of country-wide populations and are not a replacement for narrow community targeting. 

In this paper we do not attempt to solve the important problem of fairness and representation in large-scale AI models. However, our methodological framework can assist in identifying the tasks in which community-based value-alignment frameworks provide the most return on investment, which are those tasks that we identify as latent-projective content tasks. We furthermore demonstrate quantitative tools that researchers can use to evaluate the human-value or human-preference alignment of a particular ML model before it is used in production. More broadly, we hope that our recommendations will interweave with existing methods for ensuring fairness and value-alignment in AI by allowing researchers to maximize the value of resource-intensive methods and to more easily evaluate their impact.

\section{Acknowledgments}
Thank you to Aleks Yeganov for building the classifier models used in our case study. Thank you also to the anonymous AAAI reviewers for their helpful comments and feedback that improved the final version of this paper.

\bibliography{aaai25}

\begin{thebibliography}{49}
\providecommand{\natexlab}[1]{#1}

\bibitem[{Bai et~al.(2022)Bai, Kadavath, Kundu, Askell, Kernion, Jones, Chen, Goldie, Mirhoseini, McKinnon, Chen, Olsson, Olah, Hernandez, Drain, Ganguli, Li, Tran-Johnson, Perez, Kerr, Mueller, Ladish, Landau, Ndousse, Lukosuite, Lovitt, Sellitto, Elhage, Schiefer, Mercado, DasSarma, Lasenby, Larson, Ringer, Johnston, Kravec, Showk, Fort, Lanham, Telleen-Lawton, Conerly, Henighan, Hume, Bowman, Hatfield-Dodds, Mann, Amodei, Joseph, McCandlish, Brown, and Kaplan}]{bai2022constitutionalaiharmlessnessai}
Bai, Y.; Kadavath, S.; Kundu, S.; Askell, A.; Kernion, J.; Jones, A.; Chen, A.; Goldie, A.; Mirhoseini, A.; McKinnon, C.; Chen, C.; Olsson, C.; Olah, C.; Hernandez, D.; Drain, D.; Ganguli, D.; Li, D.; Tran-Johnson, E.; Perez, E.; Kerr, J.; Mueller, J.; Ladish, J.; Landau, J.; Ndousse, K.; Lukosuite, K.; Lovitt, L.; Sellitto, M.; Elhage, N.; Schiefer, N.; Mercado, N.; DasSarma, N.; Lasenby, R.; Larson, R.; Ringer, S.; Johnston, S.; Kravec, S.; Showk, S.~E.; Fort, S.; Lanham, T.; Telleen-Lawton, T.; Conerly, T.; Henighan, T.; Hume, T.; Bowman, S.~R.; Hatfield-Dodds, Z.; Mann, B.; Amodei, D.; Joseph, N.; McCandlish, S.; Brown, T.; and Kaplan, J. 2022.
\newblock Constitutional AI: Harmlessness from AI Feedback.
\newblock arXiv:2212.08073.

\bibitem[{Bakker et~al.(2022)Bakker, Chadwick, Sheahan, Tessler, Campbell-Gillingham, Balaguer, McAleese, Glaese, Aslanides, Botvinick, and Summerfield}]{bakker-22}
Bakker, M.; Chadwick, M.; Sheahan, H.; Tessler, M.; Campbell-Gillingham, L.; Balaguer, J.; McAleese, N.; Glaese, A.; Aslanides, J.; Botvinick, M.; and Summerfield, C. 2022.
\newblock Fine-tuning language models to find agreement among humans with diverse preferences.
\newblock In Koyejo, S.; Mohamed, S.; Agarwal, A.; Belgrave, D.; Cho, K.; and Oh, A., eds., \emph{Advances in Neural Information Processing Systems}, volume~35, 38176--38189. Curran Associates, Inc.

\bibitem[{Baris~Schlicht et~al.(2024)Baris~Schlicht, Altiok, Taouk, and Flek}]{baris-schlicht-etal-2024}
Baris~Schlicht, I.; Altiok, D.; Taouk, M.; and Flek, L. 2024.
\newblock Pitfalls of Conversational {LLM}s on News Debiasing.
\newblock In Hautli-Janisz, A.; Lapesa, G.; Anastasiou, L.; Gold, V.; Liddo, A.~D.; and Reed, C., eds., \emph{Proceedings of the First Workshop on Language-driven Deliberation Technology (DELITE) @ LREC-COLING 2024}, 33--38. Torino, Italia: ELRA and ICCL.

\bibitem[{Basile et~al.(2021)Basile, Fell, Fornaciari, Hovy, Paun, Plank, Poesio, and Uma}]{Basilea_21}
Basile, V.; Fell, M.; Fornaciari, T.; Hovy, D.; Paun, S.; Plank, B.; Poesio, M.; and Uma, A. 2021.
\newblock We Need to Consider Disagreement in Evaluation.
\newblock In \emph{Proceedings of the 1st Workshop on Benchmarking: Past, Present and Future}, 15--21. Association for Computational Linguistics.

\bibitem[{Basilea(2020)}]{Basilea_20}
Basilea, V. 2020.
\newblock {It’s the End of the Gold Standard as we Know it. On the Impact of Pre-aggregation on the Evaluation of Highly Subjective Tasks}.
\newblock In \emph{CEUR Workshop Proceedings}.

\bibitem[{Beigman~Klebanov and Beigman(2009)}]{beigman_09}
Beigman~Klebanov, B.; and Beigman, E. 2009.
\newblock From Annotator Agreement to Noise Models.
\newblock \emph{Computational Linguistics}, 35.

\bibitem[{Bergman et~al.(2024)Bergman, Marchal, Mellor, Mohamed, Gabriel, and Isaac}]{bergman_stela}
Bergman, S.; Marchal, N.; Mellor, J.; Mohamed, S.; Gabriel, I.; and Isaac, W. 2024.
\newblock STELA: A community-centred approach to norm elicitation for AI alignment.
\newblock \emph{Scientific Reports}, 14.

\bibitem[{Card et~al.(2020)Card, Henderson, Khandelwal, Jia, Mahowald, and Jurafsky}]{card-etal-2020-little}
Card, D.; Henderson, P.; Khandelwal, U.; Jia, R.; Mahowald, K.; and Jurafsky, D. 2020.
\newblock With Little Power Comes Great Responsibility.
\newblock In Webber, B.; Cohn, T.; He, Y.; and Liu, Y., eds., \emph{Proceedings of the 2020 Conference on Empirical Methods in Natural Language Processing (EMNLP)}, 9263--9274.

\bibitem[{Chang et~al.(2023)Chang, Rashid, Lin, Zhao, Demberg, Shi, and Chandra}]{chang_23}
Chang, E.; Rashid, M.~H.; Lin, P.-J.; Zhao, C.; Demberg, V.; Shi, Y.; and Chandra, V. 2023.
\newblock Revisiting Sample Size Determination in Natural Language Understanding.
\newblock In \emph{Findings of the Association for Computational Linguistics: ACL 2023}, 6716–6724.

\bibitem[{Chen et~al.(2024)Chen, Chen, Liu, Jiang, and Wang}]{chen2024}
Chen, G.~H.; Chen, S.; Liu, Z.; Jiang, F.; and Wang, B. 2024.
\newblock Humans or LLMs as the Judge? A Study on Judgement Biases.
\newblock arXiv:2402.10669.

\bibitem[{Christiano et~al.(2023)Christiano, Leike, Brown, Martic, Legg, and Amodei}]{christiano2023deepreinforcementlearninghuman}
Christiano, P.; Leike, J.; Brown, T.~B.; Martic, M.; Legg, S.; and Amodei, D. 2023.
\newblock Deep reinforcement learning from human preferences.
\newblock arXiv:1706.03741.

\bibitem[{Cohen(1960)}]{cohens_kappa}
Cohen, J. 1960.
\newblock A coefficient of agreement for nominal scales.
\newblock \emph{Educational and Psychological Measurement}, 20: 37--46.

\bibitem[{Fleiss and Cohen(1973)}]{fleiss_kappa}
Fleiss, J.~L.; and Cohen, J. 1973.
\newblock The equivalence of weighted kappa and the intraclass correlation coefficient as measures of reliability.
\newblock \emph{Educational and Psychological Measurement}, 33: 613--619.

\bibitem[{Fornaciari et~al.(2021)Fornaciari, Uma, Paun, Plank, Hovy, and Poesio}]{fornaciari_21}
Fornaciari, T.; Uma, A.; Paun, S.; Plank, B.; Hovy, D.; and Poesio, M. 2021.
\newblock Beyond Black {\&} White: Leveraging Annotator Disagreement via Soft-Label Multi-Task Learning.
\newblock In Toutanova, K.; Rumshisky, A.; Zettlemoyer, L.; Hakkani-Tur, D.; Beltagy, I.; Bethard, S.; Cotterell, R.; Chakraborty, T.; and Zhou, Y., eds., \emph{Proceedings of the 2021 Conference of the North American Chapter of the Association for Computational Linguistics: Human Language Technologies}, 2591--2597.

\bibitem[{Gabriel and Ghazavi(2021)}]{gabriel_valuealignment}
Gabriel, I.; and Ghazavi, V. 2021.
\newblock The Challenge of Value Alignment: from Fairer Algorithms to AI Safety.
\newblock arXiv:2101.06060.

\bibitem[{Gamer, Lemon, and Singh(2019)}]{irr_package}
Gamer, M.; Lemon, J.; and Singh, I. F.~P. 2019.
\newblock irr: Various Coefficients of Interrater Reliability and Agreement.
\newblock R package version 0.84.1.

\bibitem[{Grove et~al.(2009)Grove, Jr., Couper, Lepkowski, Singer, and Tourangeau}]{sampling_methods}
Grove, R.~M.; Jr., F. J.~F.; Couper, M.~P.; Lepkowski, J.~M.; Singer, E.; and Tourangeau, R. 2009.
\newblock \emph{Survey Methodology, 2nd Edition}.
\newblock Wiley.

\bibitem[{Hanna and Bojar(2021)}]{hanna-bojar-2021}
Hanna, M.; and Bojar, O. 2021.
\newblock A Fine-Grained Analysis of {BERTS}core.
\newblock In Barrault, L.; Bojar, O.; Bougares, F.; Chatterjee, R.; Costa-jussa, M.~R.; Federmann, C.; Fishel, M.; Fraser, A.; Freitag, M.; Graham, Y.; Grundkiewicz, R.; Guzman, P.; Haddow, B.; Huck, M.; Yepes, A.~J.; Koehn, P.; Kocmi, T.; Martins, A.; Morishita, M.; and Monz, C., eds., \emph{Proceedings of the Sixth Conference on Machine Translation}, 507--517.

\bibitem[{Howcroft et~al.(2020)Howcroft, Belz, Clinciu, Gkatzia, Hasan, Mahamood, Mille, van Miltenburg, Santhanam, and Rieser}]{howcroft-etal-2020}
Howcroft, D.~M.; Belz, A.; Clinciu, M.-A.; Gkatzia, D.; Hasan, S.~A.; Mahamood, S.; Mille, S.; van Miltenburg, E.; Santhanam, S.; and Rieser, V. 2020.
\newblock Twenty Years of Confusion in Human Evaluation: {NLG} Needs Evaluation Sheets and Standardised Definitions.
\newblock In Davis, B.; Graham, Y.; Kelleher, J.; and Sripada, Y., eds., \emph{Proceedings of the 13th International Conference on Natural Language Generation}, 169--182. Dublin, Ireland.

\bibitem[{Huang et~al.(2023)Huang, Yu, Ma, Zhong, Feng, Wang, Chen, Peng, Feng, Qin, and Liu}]{huang23}
Huang, L.; Yu, W.; Ma, W.; Zhong, W.; Feng, Z.; Wang, H.; Chen, Q.; Peng, W.; Feng, X.; Qin, B.; and Liu, T. 2023.
\newblock {A Survey on Hallucination in Large Language Models: Principles, Taxonomy, Challenges, and Open Questions}.
\newblock arXiv:2311.05232.

\bibitem[{Jamison and Gurevych(2015)}]{jamison_15}
Jamison, E.; and Gurevych, I. 2015.
\newblock Noise or additional information? Leveraging crowdsource annotation item agreement for natural language tasks.
\newblock In M{\`a}rquez, L.; Callison-Burch, C.; and Su, J., eds., \emph{Proceedings of the 2015 Conference on Empirical Methods in Natural Language Processing}, 291--297. Lisbon, Portugal.

\bibitem[{Kirk et~al.(2024)Kirk, Whitefield, Röttger, Bean, Margatina, Ciro, Mosquera, Bartolo, Williams, He, Vidgen, and Hale}]{prism_paper_24}
Kirk, H.~R.; Whitefield, A.; Röttger, P.; Bean, A.; Margatina, K.; Ciro, J.; Mosquera, R.; Bartolo, M.; Williams, A.; He, H.; Vidgen, B.; and Hale, S.~A. 2024.
\newblock The [PRISM] Alignment Project: What Participatory, Representative and Individualised Human Feedback Reveals About the Subjective and Multicultural Alignment of Large Language Models.
\newblock arXiv:2404.16019.

\bibitem[{Krippendorff(2013)}]{krippendorff_alpha}
Krippendorff, K. 2013.
\newblock \emph{Content analysis: An introduction to its methodology}.
\newblock Thousand Oaks, CA: Sage.

\bibitem[{Lavie and Agarwal(2007)}]{meteor-07}
Lavie, A.; and Agarwal, A. 2007.
\newblock METEOR: An Automatic Metric for MT Evaluation with High Levels of Correlation with Human Judgments.
\newblock In \emph{WMT@ACL}.

\bibitem[{Lin(2004)}]{rouge-04}
Lin, C.-Y. 2004.
\newblock ROUGE: A Package for Automatic Evaluation of Summaries.
\newblock In \emph{Annual Meeting of the Association for Computational Linguistics}.

\bibitem[{Lin(1991)}]{lin_91}
Lin, J. 1991.
\newblock Divergence Measures Based on the Shannon Entropy.
\newblock \emph{IEEE Transactions on Information Theory}, 37: 145 -- 151.

\bibitem[{Liu et~al.(2023)Liu, Iter, Xu, Wang, Xu, and Zhu}]{chainofthought}
Liu, Y.; Iter, D.; Xu, Y.; Wang, S.; Xu, R.; and Zhu, C. 2023.
\newblock G-Eval: NLG Evaluation using GPT-4 with Better Human Alignment.
\newblock arXiv:2303.16634.

\bibitem[{Mahaut et~al.(2024)Mahaut, Aina, Czarnowska, Hardalov, Müller, and Màrquez}]{mahaut2024}
Mahaut, M.; Aina, L.; Czarnowska, P.; Hardalov, M.; Müller, T.; and Màrquez, L. 2024.
\newblock Factual Confidence of LLMs: on Reliability and Robustness of Current Estimators.
\newblock arXiv:2406.13415.

\bibitem[{McNemar(1947)}]{mcnemars}
McNemar, Q. 1947.
\newblock Note on the sampling error of the difference between correlated proportions or percentages.
\newblock \emph{Psychometrika}, 12: 153--157.

\bibitem[{Novikova et~al.(2017)Novikova, Dusek, Curry, and Rieser}]{novikova-2017}
Novikova, J.; Dusek, O.; Curry, A.~C.; and Rieser, V. 2017.
\newblock Why We Need New Evaluation Metrics for NLG.
\newblock In \emph{Conference on Empirical Methods in Natural Language Processing}.

\bibitem[{Papineni et~al.(2002)Papineni, Roukos, Ward, and Zhu}]{bleu-02}
Papineni, K.; Roukos, S.; Ward, T.; and Zhu, W.-J. 2002.
\newblock Bleu: a Method for Automatic Evaluation of Machine Translation.
\newblock In \emph{Annual Meeting of the Association for Computational Linguistics}.

\bibitem[{Pavlovic and Poesio(2024)}]{pavlovic-poesio-2024}
Pavlovic, M.; and Poesio, M. 2024.
\newblock The Effectiveness of {LLM}s as Annotators: A Comparative Overview and Empirical Analysis of Direct Representation.
\newblock In Abercrombie, G.; Basile, V.; Bernadi, D.; Dudy, S.; Frenda, S.; Havens, L.; and Tonelli, S., eds., \emph{Proceedings of the 3rd Workshop on Perspectivist Approaches to NLP (NLPerspectives) @ LREC-COLING 2024}, 100--110. Torino, Italia: ELRA and ICCL.

\bibitem[{Perković, Drobnjak, and Botički(2024)}]{perkovic2024}
Perković, G.; Drobnjak, A.; and Botički, I. 2024.
\newblock Hallucinations in LLMs: Understanding and Addressing Challenges.
\newblock In \emph{2024 47th MIPRO ICT and Electronics Convention (MIPRO)}, 2084--2088.

\bibitem[{Peterson et~al.(2019)Peterson, Battleday, Griffiths, and Russakovsky}]{peterson_19}
Peterson, J.~C.; Battleday, R.~M.; Griffiths, T.~L.; and Russakovsky, O. 2019.
\newblock {Human uncertainty makes classification more robust}.
\newblock In \emph{IEEE/CVF International Conference on Computer Vision (ICCV)}, 9616–9625.

\bibitem[{Plank, Hovy, and S{\o}gaard(2014)}]{plank_14}
Plank, B.; Hovy, D.; and S{\o}gaard, A. 2014.
\newblock Learning part-of-speech taggers with inter-annotator agreement loss.
\newblock In Wintner, S.; Goldwater, S.; and Riezler, S., eds., \emph{Proceedings of the 14th Conference of the {E}uropean Chapter of the Association for Computational Linguistics}, 742--751. Gothenburg, Sweden.

\bibitem[{Potter and Levine‐Donnerstein(1999)}]{potter_99}
Potter, J.~W.; and Levine‐Donnerstein, D. 1999.
\newblock Rethinking validity and reliability in content analysis.
\newblock \emph{Journal of Applied Communication Research}, 27: 258--284.

\bibitem[{{R Core Team}(2013)}]{mcnemar_package}
{R Core Team}. 2013.
\newblock R: A Language and Environment for Statistical Computing.
\newblock {ISBN} 3-900051-07-0.

\bibitem[{Raina, Liusie, and Gales(2024)}]{raina2024}
Raina, V.; Liusie, A.; and Gales, M. 2024.
\newblock Is LLM-as-a-Judge Robust? Investigating Universal Adversarial Attacks on Zero-shot LLM Assessment.
\newblock arXiv:2402.14016.

\bibitem[{Reidsma and Carletta(2008)}]{reidsma_carletta_08}
Reidsma, D.; and Carletta, J. 2008.
\newblock {Reliability measurement without limits}.
\newblock \emph{Computational linguistics}, 34: 319--326.

\bibitem[{Reidsma and op~den Akker(2008)}]{reidsma_08}
Reidsma, D.; and op~den Akker, R. 2008.
\newblock {Exploiting ‘subjective’ annotations}.
\newblock In \emph{Coling 2008: Proceedings of the workshop on Human Judgements in Computational Linguistics}, 8--16.

\bibitem[{Reiter(2018)}]{reiter-2018}
Reiter, E. 2018.
\newblock {A Structured Review of the Validity of BLEU}.
\newblock \emph{Computational Linguistics}, 44(3): 393--401.

\bibitem[{Thakur et~al.(2024)Thakur, Choudhary, Ramayapally, Vaidyanathan, and Hupkes}]{thakur2024}
Thakur, A.~S.; Choudhary, K.; Ramayapally, V.~S.; Vaidyanathan, S.; and Hupkes, D. 2024.
\newblock Judging the Judges: Evaluating Alignment and Vulnerabilities in LLMs-as-Judges.
\newblock arXiv:2406.12624.

\bibitem[{Uma et~al.(2021)Uma, Fornaciari, Hovy, Paun, Plank, and Poesio}]{uma_21}
Uma, A.~N.; Fornaciari, T.; Hovy, D.; Paun, S.; Plank, B.; and Poesio, M. 2021.
\newblock Learning from Disagreement: A Survey.
\newblock \emph{Journal of Artificial Intelligence}, 72.

\bibitem[{van~der Lee et~al.(2021)van~der Lee, Gatt, van Miltenburg, and Krahmer}]{van-der-lee-etal-2020-best}
van~der Lee, C.; Gatt, A.; van Miltenburg, E.; and Krahmer, E. 2021.
\newblock Human evaluation of automatically generated text: Current trends and best practice guidelines.
\newblock \emph{Computer Speech \& Language}, 67: 1--24.

\bibitem[{van~der Lee et~al.(2019)van~der Lee, Gatt, van Miltenburg, Wubben, and Krahmer}]{van-der-lee-etal-2019-best}
van~der Lee, C.; Gatt, A.; van Miltenburg, E.; Wubben, S.; and Krahmer, E. 2019.
\newblock Best practices for the human evaluation of automatically generated text.
\newblock In \emph{Proceedings of the 12th International Conference on Natural Language Generation}, 355--368. Tokyo, Japan.

\bibitem[{Verga et~al.(2024)Verga, Hofstatter, Althammer, Su, Piktus, Arkhangorodsky, Xu, White, and Lewis}]{verga-et-al-2024}
Verga, P.; Hofstatter, S.; Althammer, S.; Su, Y.; Piktus, A.; Arkhangorodsky, A.; Xu, M.; White, N.; and Lewis, P. 2024.
\newblock Replacing Judges with Juries: Evaluating LLM Generations with a Panel of Diverse Models.
\newblock arXiv:2404.18796.

\bibitem[{Watson and Petrie(2010)}]{watson_10}
Watson, P.; and Petrie, A. 2010.
\newblock Method agreement analysis: a review of correct methodology.
\newblock \emph{Theriogenology}, 73: 1167--79.

\bibitem[{Zhang et~al.(2019)Zhang, Kishore, Wu, Weinberger, and Artzi}]{bertscore-19}
Zhang, T.; Kishore, V.; Wu, F.; Weinberger, K.~Q.; and Artzi, Y. 2019.
\newblock BERTScore: Evaluating Text Generation with BERT.
\newblock \emph{ArXiv}, abs/1904.09675.

\bibitem[{Zheng et~al.(2020)Zheng, Chiang, Sheng, Zhuang, Wu, Zhuang, Lin, Li, Li, Xing, Zhang, Gonzalez, and Stoica}]{zheng-et-al-2023}
Zheng, L.; Chiang, W.-L.~C.; Sheng, Y.; Zhuang, S.; Wu, Z.; Zhuang, Y.; Lin, Z.; Li, Z.; Li, D.; Xing, E.~P.; Zhang, H.; Gonzalez, J.; and Stoica, I. 2020.
\newblock Judging LLM-as-a-judge with MT-Bench and Chatbot Arena.
\newblock In \emph{Proceedings of the 37th International Conference on Neural Information Processing Systems}, NIPS '23.

\end{thebibliography}

\end{document}